\documentclass[runningheads]{llncs}

\usepackage[mobile]{eccv}

\usepackage{eccvabbrv}

\usepackage{graphicx}
\usepackage{booktabs}
\usepackage{bm}
\usepackage{amsmath}
\usepackage{wrapfig}

\DeclareMathOperator*{\argmax}{arg\,max} 

\usepackage[accsupp]{axessibility}  

\usepackage{hyperref}

\usepackage{orcidlink}

\usepackage{array} 
\newcolumntype{C}[1]{>{\centering\arraybackslash}p{#1}}

\providecommand{\thetitle}{}
\let\oldtitle\title
\renewcommand{\title}[1]{\oldtitle{#1}\renewcommand{\thetitle}{#1}}
\newcommand{\maketitlesupplementary}{
    \newpage
    \begin{center}
        \Large
        \textbf{\thetitle}\\[0.5em] 
        Supplementary Material\\[1.0em]
    \end{center}
}

\usepackage{listings}
\usepackage{xcolor}
\usepackage{amsmath}
\usepackage{amsfonts}
\usepackage{amssymb}

\definecolor{codegreen}{rgb}{0,0.6,0}
\definecolor{codegray}{rgb}{0.5,0.5,0.5}
\definecolor{codepurple}{rgb}{0.58,0,0.82}
\definecolor{backcolour}{rgb}{0.95,0.95,0.92}

\lstdefinestyle{mystyle}{
    backgroundcolor=\color{backcolour},   
    commentstyle=\color{codegreen},
    keywordstyle=\color{magenta},
    numberstyle=\tiny\color{codegray},
    stringstyle=\color{codepurple},
    basicstyle=\footnotesize\ttfamily,
    breakatwhitespace=false,         
    breaklines=true,                 
    captionpos=b,                    
    keepspaces=true,                 
    numbers=left,                    
    numbersep=5pt,                  
    showspaces=false,                
    showstringspaces=false,
    showtabs=false,                  
    tabsize=2,
    language=Python
}
\begin{document}


\title{FlashSplat: 2D to 3D Gaussian Splatting Segmentation Solved Optimally} 


\titlerunning{FlashSplat}


\author{ Qiuhong Shen  \and 
         Xingyi Yang  \and 
         Xinchao Wang\thanks{Corresponding Author.}\orcidlink{0000-0003-0057-1404}
}

\authorrunning{Shen et al.}

\institute{National University of Singapore \\
\email{\{qiuhong.shen,xyang\}@u.nus.edu} \quad
\email{xinchao@nus.edu.sg}}





\maketitle
\begin{abstract}

This study addresses the challenge of accurately segmenting 3D Gaussian Splatting~(3D-GS) from 2D masks. Conventional methods often rely on iterative gradient descent to assign each Gaussian a unique label, leading to lengthy optimization and sub-optimal solutions. Instead, we propose a straightforward yet globally optimal solver for 3D-GS segmentation. The core insight of our method is that, with a reconstructed 3D-GS scene , the rendering of the 2D masks is essentially a linear function with respect to the labels of each Gaussian. As such, the optimal label assignment can be solved via linear programming in closed form. This solution capitalizes on the alpha blending characteristic of the splatting process for single step optimization. By incorporating the background bias in our objective function, our method shows superior robustness in 3D segmentation against noises. Remarkably, our optimization completes within 30 seconds, about 50$\times$ faster than the best existing methods. Extensive experiments demonstrate our method’s efficiency and
robustness in segmenting various scenes, and its superior performance in
downstream tasks such as object removal and inpainting. Demos and code will be available at \href{https://github.com/florinshen/FlashSplat}{https://github.com/florinshen/FlashSplat}.

\keywords{3D Segmentation \and 3D Gaussian Splatting \and Neural-based understanding}
\end{abstract}
\section{Introduction}

The pursuit of understanding and interacting with 3D environments represents a formidable yet pivotal challenge in computer vision. Central to this endeavor is the task of accurately perceiving and segmenting 3D structures~\cite{NeRF, dvgo, tensorf, mipnerf360, InstantNGP, plenoxel}, a task that becomes increasingly complex as we delve into more sophisticated representations of 3D scenes. 
Recently, 3D Gaussian Splatting (3D-GS)~\cite{3dgs} emerges as a cutting-edge approach that promises to revolutionize how we render and reconstruct 3D spaces. By employing a multitude of colored 3D Gaussians, this method achieves a high fidelity representation of 3D scenes, offering a compelling blend of precision and visual quality that is particularly suited for complex object and scene rendering.

Despite the potential of 3D-GS, a significant obstacle remains: the segmentation of these 3D Gaussian from mere 2D masks—a process crucial for a range of applications from object recognition to scene manipulation. Existing approaches~\cite{gsgroup, saga} to this challenge have largely depended on iterative gradient descent methods for labeling 3D Gaussian. However, these methods are marred by their slow convergence speed and frequent entrapment in suboptimal solutions, rendering them impractical for applications requiring real-time performance or high accuracy.

Addressing this gap, our work introduces a simple but globally optimal solver designed explicitly for the segmentation of 3D Gaussian Splatting. Our approach capitalizes on the insight that the process of rendering segmented 2D images from a reconstructed 3D Gaussian Splatting can be simplified to a linear function concerning the accumulated contribution of each Gaussian. This realization allows us to frame the problem as a linear integer programming that can be solved in a closed form, relying solely on the alpha composition term inherent in the splatting process. This breakthrough significantly streamlines the segmentation task, bypassing the need for iterative optimization and directly leading to the optimal label assignment.

Moreover, by integrating a background bias into our objective function, we further enhance the method's robustness against 2D masks noise in 3D segmentation. This refinement not only bolsters the robustness of our solution but also its applicability across a wider range of scenes and conditions. Impressively, our solver completes its optimization in approximately 30 seconds—an acceleration of 50$\times$ faster than existing methods—without sacrificing accuracy or global optimality.

Through extensive experimentation, we have validated the superiority of our approach in efficiently segmenting a variety of scenes, demonstrating its enhanced performance in downstream tasks such as object removal and inpainting. These results underscore the potential of our method to significantly advance the field of 3D scene processing and understanding.

The primary contributions of our work are summarized as follows:
\begin{itemize}
    \item We introduce a globally optimal solver for 3D Gaussian
Splatting segmentation, which significantly enhances the efficiency of lifting 2D segmentation results into 3D space.
    \item We simplify the 3D-GS rendering process through linearization, turning the 2D to 3D segmentation task into a linear integer optimization problem. This method is effective for both binary and scene segmentation.
    \item We introduce a background bias in the optimization, demonstrate superior robustness againt noises in 3D segmentation, showcasing our method's robustness and efficiency across a variety of scene segmentation.
    \item Our method achieves a remarkable optimization speed, completing the process within 30 seconds—approximately 50 times faster than existing methods—while ensuring global optimality for given 2D masks.
    \item Extensive experiments validate the superiority of our approach in downstream tasks, including object removal and inpainting, thus highlighting its potential to significantly impact 3D data processing and application.
\end{itemize}

\label{sec:intro}

\section{Related-work}
\label{sec:related-work}

\subsection{3D Gaussian Splatting} 3D Gaussian Splatting has emerged as an efficient method for inverse rendering, facilitating the reconstruction of 3D scenes from 2D images through learning explicit 3D Gaussian in the space~\cite{kerbl20233d, 3dgs_survey}. Recent advancements have seen a series of studies~\cite{yang2023real, luiten2023dynamic, yang2023deformable, gflow} extending this approach to capture dynamic 3D environments by learning deformation fields for 3D Gaussians. Concurrently, another stream of researches~\cite{tang2023dreamgaussian, chen2023text, yi2023gaussiandreamer, aligngs, dreamgaussian4d, 4dgen} have integrated 3D Gaussian splatting with diffusion-based models~\cite{zero123, stable-diffusion, anything3d, hash3d, consistent3d, dtc123} to generate static~\cite{tgs, gamba, mvgamba} or dynamic~\cite{l4gm} 3D objects. A notable portion of these works leverages the explicit representations provided by 3D Gaussian Splatting for rapid optimization. Our method primarily concentrates on elevating corresponding 2D masks onto reconstructed 3D Gaussian scenes. We also exploit the explicit representation of 3D Gaussian Splatting to cast the segmentation of 3D scenes as a Linear Programming optimization problem, thereby enabling instantaneous 3D segmentation.



\subsection{3D Neural scene segmentation} 
Recent advancements in neural 3D scene representations, including NeRF, DVGO, and several others~\cite{NeRF, dvgo, tensorf, mipnerf360, InstantNGP, bakingnerf, plenoxel, nex, autoint, TiNeuVox}, have demonstrated exceptional capabilities in 3D scene reconstruction. Inspired by these developments, the exploration of 3D segmentation within such frameworks has emerged as a significant area of interest. Semantic-NeRF~\cite{semantic-nerf} initially showcased NeRF's aptitude for semantic propagation and refinement. NVOS~\cite{nvos} furthered this by enabling user interaction for 3D object selection via a lightweight MLP trained on specialized 3D features. Concurrently, a series of studies including N3F, DFF, LERF, and ISRF~\cite{n3f, dff, lerf, isrf} have integrated 2D vision models to facilitate 3D segmentation by elevating 2D visual features, thereby processing embedded 3D features for segmentation. Additionally, various approaches have combined instance and semantic segmentation methods with radiance fields~\cite{obsurf,giraffe,uorf,rfp, instance-nerf, dmnerf, panopticnerf, nesf, sa3d, isrf} to further explore this domain.
%
%

Most closely related methods to our work are SAGA~\cite{saga}, Gaussian Grouping~\cite{gsgroup}, and SAGS~\cite{sa3g}. 
These works concentrate on associating 2D masks, as generated by the Segment Anything Model (SAM)~\cite{sam}, with 3D Gaussian Splatting for mask lifting. Gaussian Grouping employs video tracker to associate 2D masks, subsequently distilling these as object features for 3D Gaussians and segmenting via a classifier trained with 2D identity loss and 3D regularization loss. Similarly, SAGA distills 2D masks into 3D features, requiring additional training features for each 3D Gaussians with a pack of losses. Both of these methods are limited by their substantial training costs required to optimize Gaussian features before 3D segmentation. SAGS introduces a training-free approach by projecting the centers of 3D Gaussians onto 2D masks. This method adopts a intuitive segmentation criterion, where a projected center landing within a foreground mask gets classified as a foreground Gaussian.
 
Our method distinguishes itself by framing the 2D mask lifting in 3D Gaussian Splatting as a Linear Programming (LP) optimization problem, solved in a single step. Focused solely on lifting associated 2D masks into 3D Gaussians masks, our approach offers a standalone, efficient solution for immediate segmentation without the need for extra training and post-processing.

\section{Methods}



In this section, we first delve into the rendering process of 3D Gaussian Splatting (3DGS), focusing on the tile-based rasterization and alpha blending. We then describe how this process lends itself to formulating the segmentation of 3DGS as an integer linear programming (ILP) optimization, which we demonstrate can be resolved in a closed form. Recognizing the typically noisy nature of 2D masks, we introduce a softened optimal assignment to mitigate these noises. 
Beyond binary segmentation, we extend our method to include scene segmentation, enabling the segmentation of all objects within 3D scenes. Finally, we present a method for rendering 2D masks guided by depth information, which projects the 3D segmentation results onto 2D masks from new viewpoints.

\subsection{Preliminary: Rasterization of 3DGS}

\begin{wrapfigure}{r}{0.4\textwidth}
\vspace{-4mm}
  \centering
  \includegraphics[width=0.4\textwidth]{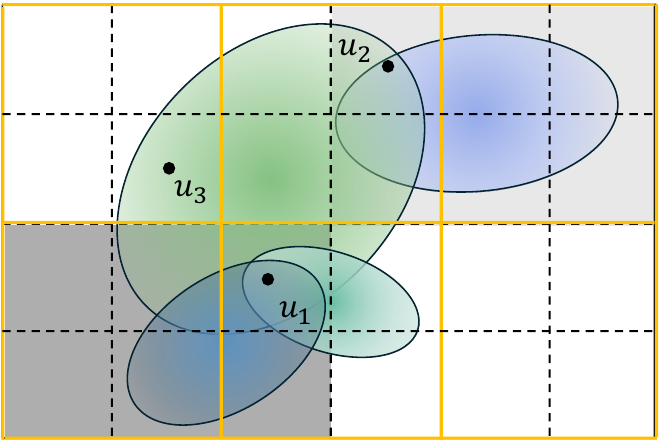}
  \caption{\textbf{GS Rasterization.} Here we illustrate projected 2D Gaussians in 3DGS rasterization, setting each tile as $2\times 2$ for illustration purposes. Gaussians are shared between different tiles and are likely to be shared among different instances (color blocks). }
  \label{fig:gsrender}
  \vspace{-4mm}
\end{wrapfigure}

3D Gaussian Splatting~\cite{kerbl20233d} (3DGS) stands out as a great method for novel view synthesis. Unlike neural radiance fields~\cite{mildenhall2020nerf}, it reconstructs a 3D scene as explicit 3D Gaussians with real-time speed. Given a set of captured views with paired camera poses, 3D Gaussian Splatting reconstruct 3D scenes by learning 3D Gaussians $\{G_{i}\}$. Each 3D Gaussian $G_{i}$ is parameterized as $G_{i} = \{m_i, q_i, s_i, o_i, c_i\}$, where $m_i \in \mathbb{R}^3$ is the center position, $q_i \in \mathbb{R}^4$  is a quaternion representing rotations, $s_i \in \mathbb{R}^3$ is the scale of three dimensions, $o_i \in \mathbb{R}$ is the learnt opacity, and $c_i \in \mathbb{R}^{48}$ is the three-order spherical harmonics to represent view dependent color. In its rendering process, 3DGS adopts a rasterization pipeline to achieve super efficiency. Specifically, all 3D Gaussians are first projected onto the image plane as 2D Gaussians. Then the whole image is divided into $16\times 16$ tiles as illustrated in Fig.~\ref{fig:gsrender}, pixels in each tile $B$ share identical 3D Gaussians subset $\{G_{i}\}_{B} \in \{G_{i}\} $. When rendering each pixels, the traditional alpha composition is applied to blend properties $x_i$ (color or depth) of these 2D Gaussians into pixel space property $X$ by their depth order:

\begin{equation}
X = \sum_{i \in \{G_{i}\}_{B}} x_i \alpha_{i} \prod_{j=1}^{i-1}(1 - \alpha_{j}) = \sum_{i \in \{G_{i}\}_{B}} x_i \alpha_{i} T_i
\label{eq:alpha_blend}
\end{equation}
where $\alpha_{i}$ is the alpha value when rendering specific pixels, which is the product of  $o_i$ and the probability of pixel position drops in the projected 2D Gaussians parameterized as $(m^{2D}_i, \Sigma^{2D}_i)$,  and $T_i = \prod_{j=1}^{i-1}(1 - \alpha_{j})$ is the transmittance, presenting the fraction that passes through and is not absorbed by the preceding $i - 1$ Gaussians.

\vspace{-4mm}
\subsection{Binary Segmentation as Integer Linear Programming}

We start with a reconstructed 3DGS scene, parameterized by $\{G_{i}\}$. It includes $L$ rendered views with associated 2D binary masks, denoted as $\{M^v\}$. Each mask, $M \in \mathbb{R}^{H \times W}$ has elements of 0 or 1, where 0 represents the background, and 1 denotes the foreground. Our goal is to assign a 3D label, $P_i$, which can be either 0 or 1, to each 3D Gaussian $G_i$ by projecting the 2D masks into the 3D space.

To do this, we optimize $P_{i}$ through a differentiable rendering process. 
We assume $P_{i}$ as a property $x_i$ in Eq.~\ref{eq:alpha_blend}. The aim is to minimize the discrepancy between the rendered masks and the provided binary masks ${M^v}$. 

Luckily, with $\{G_{i}\}$ reconstructed, the $\alpha_i$ and the $T_i$ become constant for rendering. As such, the rendering function becomes a linear equation with respect to the blending properties $x_i$. This gives us a great flexibility to solve for the optimal mask assignment using simple linear optimization.


Formally, our segmentation problem can be formulated as an integer Linear Programming (LP) optimization with mean absolute error:
\begin{equation}
\begin{aligned}
& \underset{\{P_{i}\}}{\text{Min}}
& & \mathcal{F} = \sum_{v} |\mathcal{R}(\{G_{i} \}, \{P_{i}\}) - M^v|= \sum_{v \in L} \sum_{M_{jk}^{v} \in M^v} \left| \sum_{i} P_{i} \alpha_{i} T_{i} - M_{jk}^{v} \right| \\
& \text{subject to}
& & P_{i} \in \{0, 1\}.
\end{aligned}
\label{eq:optimal_obj}
\end{equation}
where $\mathcal{R}$ denotes the 3DGS renderer. Given $\alpha_{i} \geq 0$ and $T_{i} \leq 1$ for any $i$, and the given light in alpha composition can only be absorbed, the total absorbed light cannot exceed the initial light intensity, which is normalized to 1. Consequently, the sum of absorbed light fractions across all samples is bounded by $ 0< \sum_{i}\alpha_{i}T_{i} \leq 1$. This leads us to introduce:

\noindent\textbf{Lemma 1:} \textit{In the alpha blending within 3D Gaussian Splatting,  for $P_{i} \in \{0, 1\}, $}

\begin{equation}
0 \leq \sum P_{i} \alpha_{i} T_{i} \leq \sum \alpha_{i} T_{i} \leq 1
\end{equation}

To solve this LP, we refer \textit{Lemma 1} to rewrite the Eq~\ref{eq:optimal_obj} as:
\begin{align}
\min \mathcal{F} &= \sum_{v,i,j,k} P_{i} \alpha_{i} T_{i} \mathbb{I}(M_{jk}^{v}, 0) - \sum_{v,i,j,k} P_{i} \alpha_{i} T_{i} \mathbb{I}(M_{jk}^{v}, 1) + C \label{eq:binary_1} \\
&= C + \sum_{i}P_{i} (\sum_{v,j,k} \alpha_{i} T_{i} \mathbb{I}(M_{jk}^{v}, 0) - \sum_{v,j,k} \alpha_{i} T_{i} \mathbb{I}(M_{jk}^{v}, 1)) \label{eq:binary_2} \\
&= C + \sum_{i} P_{i} (A_{0}^{i} - A_{1}^{i}) \label{eq:binary_3}
\end{align}
where $C = \sum_{v,j,k} M_{jk}^{v}$ is a constant value, alongside $\mathbb{I}(\cdot, 0)$ and $\mathbb{I}(\cdot, 1)$, serve as indicator functions signifying background and foreground presence, respectively.

\noindent\textbf{Solving ILP as Majority Vote.} To minimize the function $\mathcal{F}$, as outlined in Eq.~\ref{eq:binary_1}, we assign $P_i = 0$ when all the corresponding pixels in masks ${M^v}$ indicate background (0). Conversely, if there is a contradiction among the masks, we resolve this by as a weighted majority vote. Specifically, we assign the value of each Gaussian $G_i$ based on the most frequent label in the masks, as detailed in Eq~\ref{eq:binary_3}.

Formally, if the difference $(A_{0} - A_{1}) > 0$, where $A_0$ and $A_1$ represent the counts of weighted masks indicating background and foreground respectively, we assign $P_i = 0$ to minimize $\mathcal{F}$. If $(A_{0} - A_{1}) < 0$, then $P_i = 1$ is assigned. In a more general context, we can formulate the optimal assignment for $\{P_i\}$ as follows:
\begin{equation}
\begin{aligned}
    P_{i} &= \argmax_{n} A_{n}, \; n \in \{0, 1\}\\
    \text{where} \;\; A_{n} &= \sum_{v,j,k} \alpha_{i} T_{i} \mathbb{I}(M_{jk}^{v}, n)
\end{aligned}
\label{eq:optimal_assignment}
\end{equation}
Intuitively, this optimal assignment is conceptualized to aggregate the contributions of individual 3D Gaussians during its rendering. Gaussians that significantly contribute to the foreground in all given masks are designated for foreground as $P_{i} = 1$, conversely, those contributing predominantly to the background are systematically allocated to the background as $P_{i} = 0$. Additionaly, thanks to the simple linear combination form in the objective of Equation~\ref{eq:binary_3}, we can concurrently assign optimal labels to each 3D Gaussians. 

\begin{figure}[h!]
\centering
\includegraphics[width=0.9\linewidth]{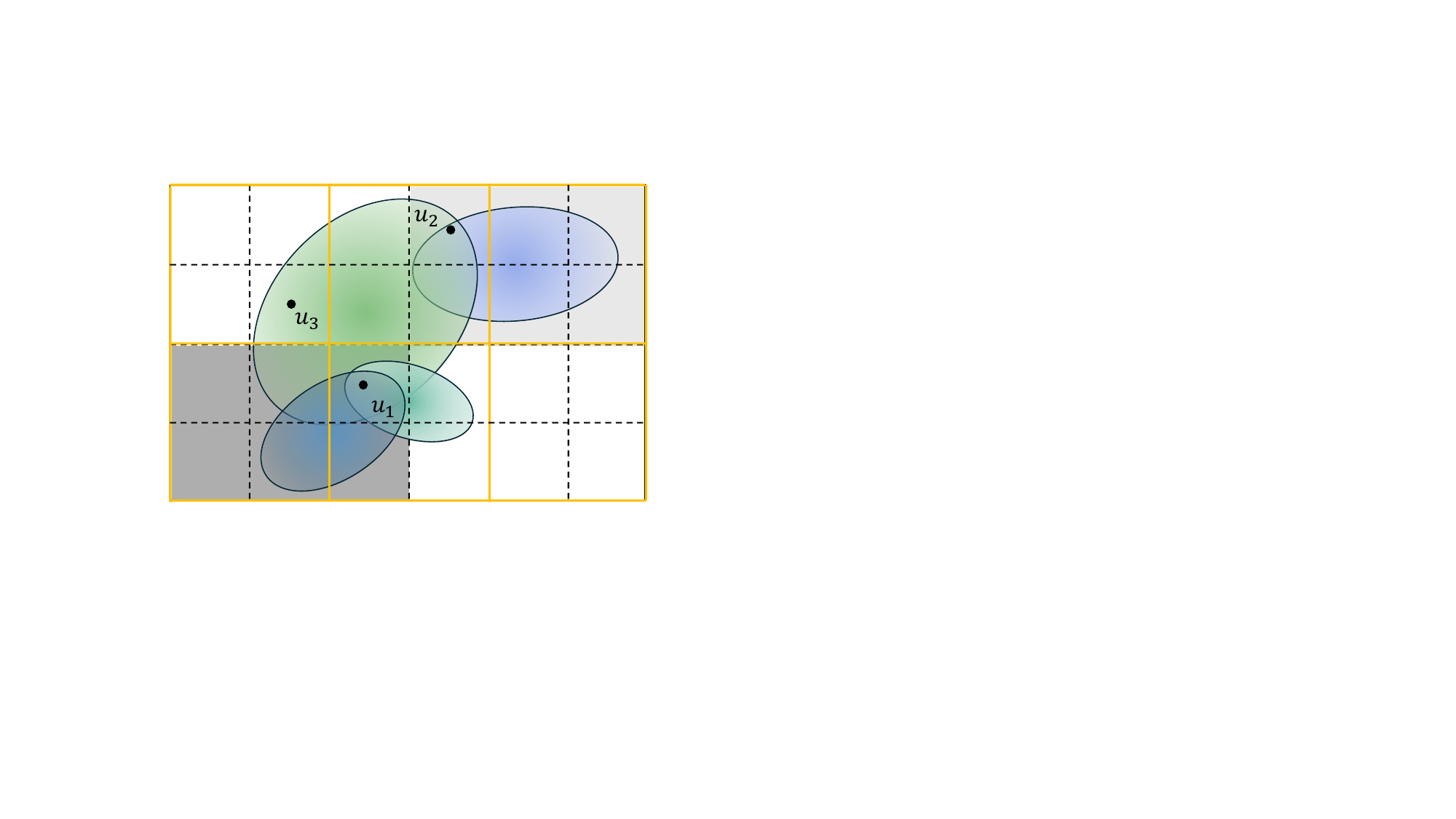}
\caption{\textbf{The effect of the background bias $\gamma$}. By adjusting the background bias in our optimal assignment, the noisy 3D segmentation caused by 2D masks can be flexibly mitigated for various downstream applications.}
\label{fig:relaxtion_factor}
\vspace{-6mm}
\end{figure}

\noindent\textbf{Regularized ILP for Noise Reduction.} Practically, the given 2D mask set $\{M^v\}$ is typically predicted and associated by trained 2D vision models, which is likely to introduce noise in certain regions (illustrated in column 1 and column 2 of Fig.~\ref{fig:noise_reduction}). This characteristic of the provided 2D masks can lead to noisy 3D segmentation results. For instance, as depicted in Fig.\ref{fig:relaxtion_factor} (a), background Gaussians misclassified as foreground due to noise can result in a segmented 3D object with sharp, difficult-to-filter edges. 

To address this, we refine above optimal assignment. This refinement entails initially normalize overall contributions of Gaussians through $L1$ normalization, represented as $\bar{A_{e}} = A_{e} / \sum_{t} A_{t}$. We then introduce a background bias, ranging between $\gamma \in [-1, 1]$,  to recalibrate $\hat{A_{0}} = \bar{A_{0}} + \gamma$, thereby adjusting the optimal assignment as $P_{i} = \argmax_{n} \{\hat{A_{0}}, \bar{A_{1}}\}$. Employing of $\gamma > 0$ effectively diminish foreground noise in segmentation outcomes; conversely, $\gamma < 0$ results in a cleaner background, as demonstrated in Fig.\ref{fig:relaxtion_factor} (b) and (c), respectively. This softened form of optimal assignment provides flexibility to produce accurate 3D segmentation results against 2D masks noises for different downstream tasks.

\subsection{From Binary to Scene Segmentation}
\label{subsec:multi-seg}
Numerous instances are present across various views in 3D scenes. Segmenting multiple objects within these scenes requires executing binary segmentation multiple times according to above formulation. This process involves repeatedly gathering $\{A_{0}, A_{1}\}$, which inherently slows down the pace of scene segmentation. Thus, we extend our methodology from binary segmentation to encompass scene segmentation to address this challenge more efficiently.

This transition to multi-instance segmentation is motivated by two key considerations. Initially, it's important to recognize that 3D Gaussians do not exclusively belong to a single object. This is exemplified in Figure~\ref{fig:gsrender}, where pixel $u_{1}$ and $u_{2}$ are influenced by the same 3D Gaussian, despite belonging to distinct objects (color block). Furthermore, the introduction of multiple instances complicates the constraint in Equation~\ref{eq:optimal_obj} to $P_{i} \in \{0, 1, ..., E - 1\}$, with $E$ representing the total number of instances in a scene. Consequently, the set of provided masks ${M^v}$ becomes $M^v \in \{0, 1, ..., E-1\}$ to accommodate multiple segmented 2D instances. Such constraints prevent achieving a global optimum, as the labels across instances are subject to be exchangable.

To circumvent these challenges, we reinterpret multi-instance segmentation as a combination of binary segmentation, modifying the optimal assignment strategy as outlined in Equation~\ref{eq:optimal_assignment}. To isolate a specific instance labeled $e$ in a 3D scene, we redefine all other objects within the instance set as the background to compute $A_{others}$. This approach is formalized as follows:
\begin{equation}
\begin{aligned}
    P_{i} &= \argmax_{n} A_{n}, \; n \in \{0, t\}\\
    \text{where} \;\; A_{t} &= \sum_{v,j,k} \alpha_{i} T_{i} \mathbb{I}(M_{jk}^{v}, t), \\
    A_{0} &= A_{others}= \sum_{e \neq t} \sum_{v,j,k} \alpha_{i} T_{i} \mathbb{I}(M_{jk}^{v}, e)
\end{aligned}
\label{eq:multi_assignment}
\end{equation}
With this formulation, we only need to accumulate the set $\{A_{e}\}$ once, and then perform $\argmax$ on this set to get Gaussians subset $\{G_i\}_{e}$ for each object $e$. Consequently, this allows users to selectively remove or modify these 3D Gaussian subsets by specifying object ID. It is also noteworthy that subsets $\{G_i\}_{e}$ from different instances may overlap under this formulation when $\gamma < 0$, a reflection of the inherent non-exclusive nature of 3D Gaussian Splatting (see more details in supplementary material Sec. 8.3).

\subsection{Depth-guided Novel View Mask Rendering}
\label{sec:mask_render}
Given above formulation eschews the need for dense optimization, our approach is capable of yielding robust segmentation results using merely approximately 10\% of the masked 2D views. This efficiency also empowers our method with the capability to produce 2D masks $\hat{M}^{v}$ for previously unseen views. For rendering masks in novel views within the context of binary segmentation, we focus exclusively on rendering foreground Gaussians identified by $P_{i} = 1$. This process involves computing an accumulated alpha value $\rho_{jk}$ for every pixel $(j, k)$. For novel view mask rendering in binary segmentation, we merely render foreground Gaussians with $P_{i} = 1$ to produce accumulated alpha value $\rho_{jk}$ for each pixel, then the 2D mask can be obtained by simple quantization on alpha values $\rho_{jk}$ with threshold $\tau$, which is a pre-defined hyper parameter. Specifically, 2D masks are derived through straightforward quantization of these alpha values, where $\hat{M}^{v}_{jk} = \mathbb{Q}(\rho_{jk}, \tau)$ and $\tau$ represents a predetermined threshold, and $\mathbb{Q}$ is the quantization function, when $\rho_{jk} > \tau$, $\mathbb{Q} = 1$, conversely $\mathbb{Q}=0$.

\begin{figure}[h!]
\vspace{-4mm}
\centering
\includegraphics[width=1.0\linewidth]{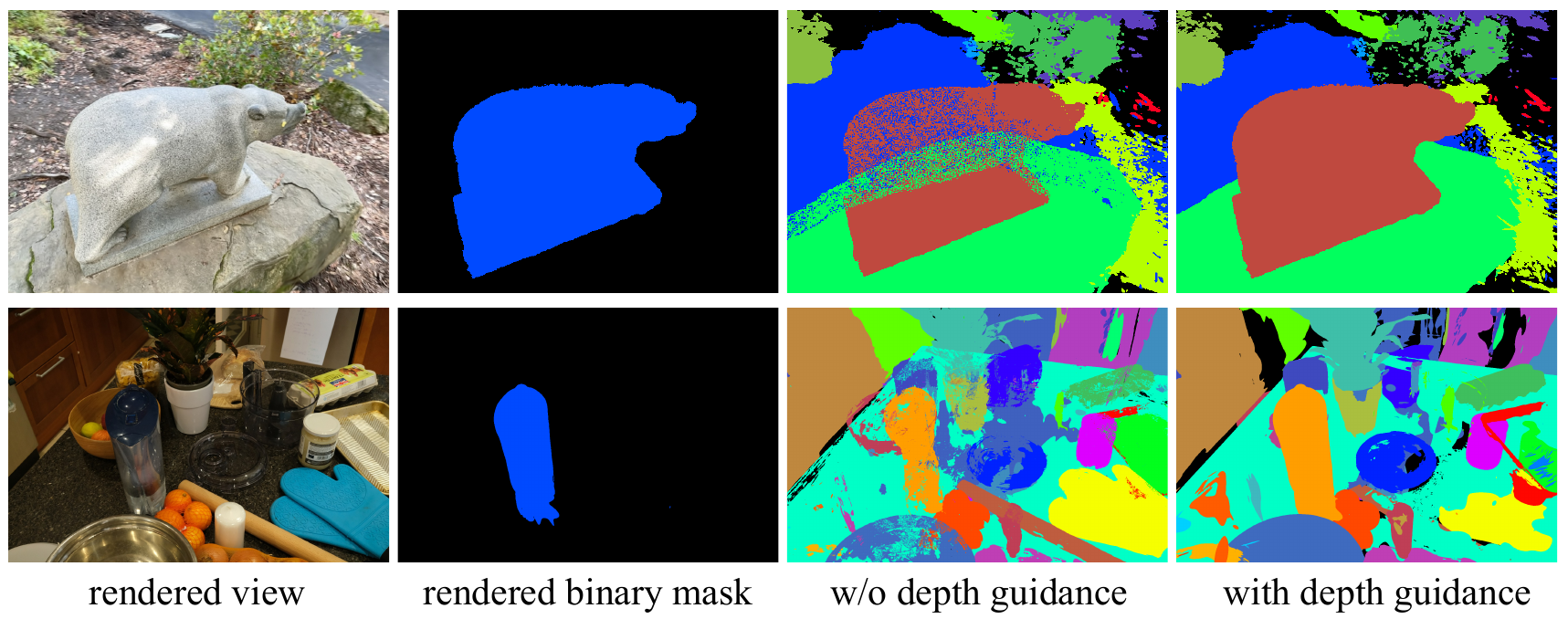}
\caption{\textbf{Novel view mask rendering}. Here we showcase mask rendering on novel views for both binary and scene segmentation. Simple alpha mask quantization can generate consistent masks in binary segmentation. With depth guidance, scene segmentation also can generate feasible 2D masks in novel views.}
\label{fig:depth_guided_mask}
\vspace{-6mm}
\end{figure}

For novel view mask rendering in scene segmentation, due to the intersections among segmentation results of different objects as indicated in Equation~\ref{eq:multi_assignment}, the resultant alpha mask might present ambiguity, as illustrated in 3rd column of Fig.~\ref{fig:depth_guided_mask}. Specifically, rendering each object's associated 3D Gaussian subset $\{G_i\}_e$ in the same viewpoint might result in several objects meeting the condition $\mathbb{Q}( \rho_{jk}^{e}, \tau) = 1$. In this case, we introduce depth to determine the final segmentation results. The depth at each pixel location $(j, k)$ is utilized to filter the final 2D mask outcome. The object $e$ satisfying $\mathbb{Q}( \rho_{jk}^{e}, \tau) = 1$ with minimal depth $D_{jk}^{e}$ relative to the camera at a given pixel $(j, k)$ is selected as $\hat{M}_{jk} = e$. Figure~\ref{fig:depth_guided_mask} illustrates the prediction of 2D masks for both binary and multi-object segmentation in novel views.
\label{sec:methods}
\section{Experiments}
\label{sec:experiments}

\subsection{Data preparation}
\textbf{Dataset.} To assess the efficacy of our approach, we collect 3D scenes from several sources: the MIP-360~\cite{mipnerf360} dataset, T\&T~\cite{tanks} dataset, LLFF~\cite{llff} dataset, Instruct-NeRF2NerF~\cite{instructnerf}, and the LERF~\cite{lerf} dataset, serving as the basis for qualitative analysis. For quantitative analysis, we utilize the NVOS~\cite{nvos} dataset.
\vspace{2mm}

\noindent{\textbf{2D mask generation and association.}} In our experimental setup, we utilize the Segment Anything Models (SAM)~\cite{sam} to extract masks, given that SAM's segmentation output is inherently semantic-agnostic. It becomes necessary to further associate these 2D masks within our framework. Our approach diverges into two distinct strategies, tailored respectively for binary and scene segmentation. For binary segmentation, where the objective is to isolate a single foreground entity, we initiate by marking point prompts on a single reference view. These point prompts are projected back to the 3D space with reference view camera pose to find their nearest 3D Gaussians with the smallest positive depth. Subsequently, these point prompts are propagated to other views by projecting their corresponding 3D Gaussians' center. Leveraging these associated point prompts, SAM independently generates a binary mask for each view. For of scene segmentation, our methodology begins with employing SAM to produce instance masks for individual views. To assign each 2D object with a unique ID in the 3D scene, multiple views are treated akin to a video sequence. Utilizing a zero-shot video tracker~\cite{track-anything, gsgroup}, we ensure the consistent association and propagation of objects across viewpoints.

\subsection{Implementation details} 
We implement the optimal 3D segmentation both in Eq.~\ref{eq:optimal_assignment} and Eq.~\ref{eq:multi_assignment} within CUDA kernel functions. The computation of $A_{i}$ for each Gaussian employs tile-based rasterization with the renderer, excluding the color component $c_i$ and focusing solely on $G_{i} = \{m_i, q_i, s_i, o_i\}$. For each pixel in the provided mask set ${M}$, we perform alpha composition as defined in Equation~\ref{eq:alpha_blend}. The product $\alpha_{i}T_{i}$ for each Gaussian is then aggregated into the buffer $A_{e}$, corresponding to the pixel label $e = M^v_{jk}$, through atomic operations. Leveraging the rasterization pipeline enables the completion of the set ${A_{e}}$ for all objects in under 30 seconds. For binary segmentation, the $\argmax$  is directly applied to determine the optimal assignment. In the case of scene segmentation, $\argmax$ is executed iteratively alongside dynamic programming to speedup. This assignment ensures that the computation for both binary and scene segmentation concludes in 1 millisecond, thereby facilitating interactive adjustment of the $\gamma$ value once the set $\{A_{e}\}$ is computed. This adjustment allows users to effectively reduce segmentation noise across a variety of downstream tasks.

\subsection{3D segmentation results}
\begin{figure}[h!]
\vspace{-4mm}
\centering
\includegraphics[width=1.0\linewidth]{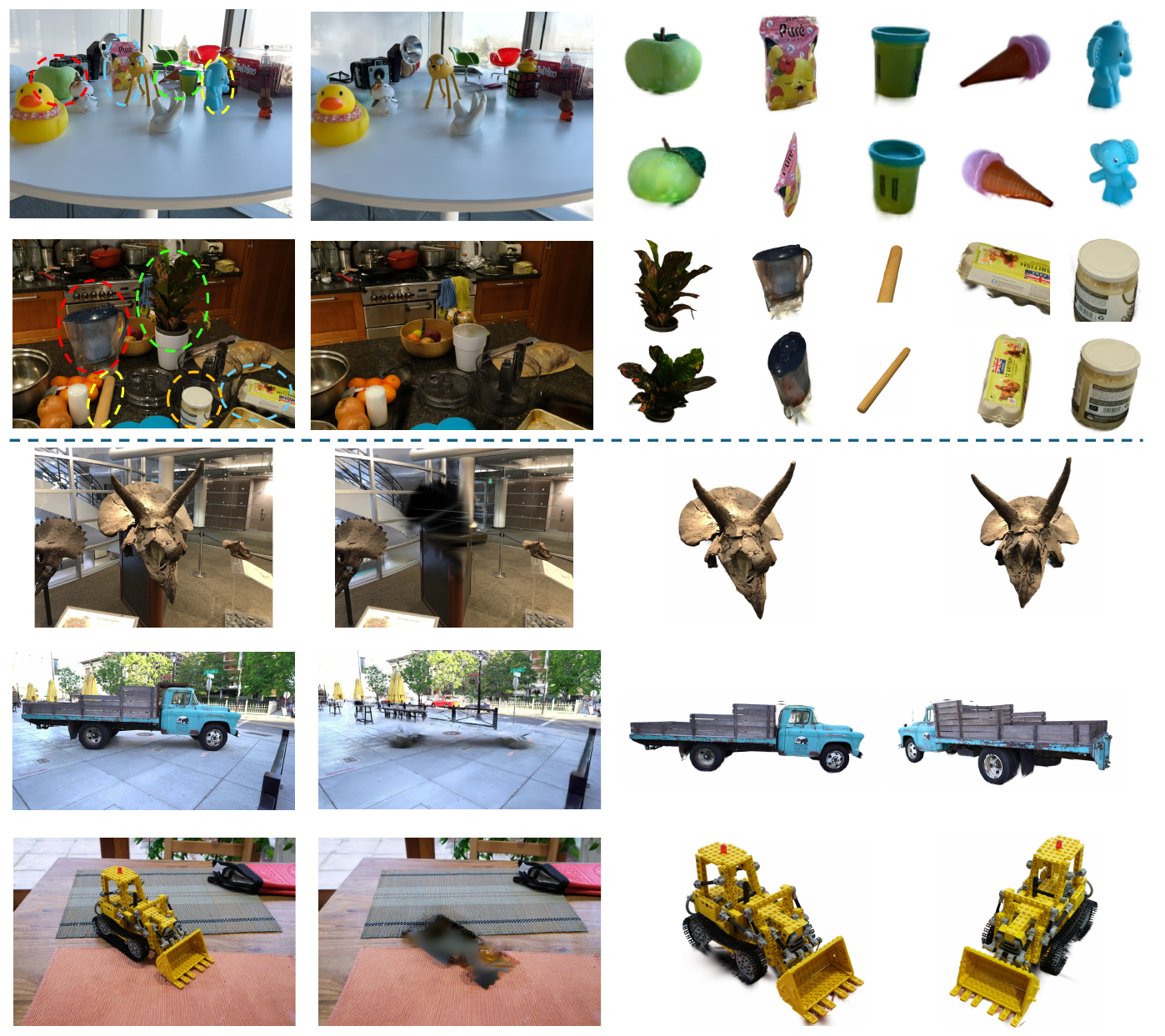}
\caption{\textbf{Qualitative result of FlashSplat.} FlashSplat is capable of performing both binary segmentation and scene segmentation. With our single step optimal assignment, all 3D segmentation is completed within 30 seconds. These segmentation results show our FlashSplat can robustly segment 3D objects and remove objects in 3D scenes.}
\label{fig:3d_segmentation}
\vspace{-6mm}
\end{figure}

In Fig.\ref{fig:3d_segmentation}, we exhibit the results of both binary and scene 3D segmentation. The first row presents the Figurines scene from the LERF dataset\cite{lerf}, and the second row features the Counter scene from the MIP-360 dataset~\cite{mipnerf360}. On both scenes, we apply our scene segmentation approach, rendering 2 views for 5 segmented objects (circled in the ground-truth images) for each scene, demonstrating our method's capability in conducting scene segmentation with instance masks predicted by SAM~\cite{sam}. Additionally, binary segmentation is showcased in rows 3, 4 and 5, with row 3 illustrating the Horns scene from the LLFF dataset~\cite{llff}, row 4 displaying the Truck scene from the T\&T dataset~\cite{tanks}, row 5 displaying the Kitchen scene from the MIP-360 dataset~\cite{mipnerf360}. Two views of the segmented objects are rendered, showing our approach's capability in segmenting 3D objects.

\subsection{Object Removal}
\label{sec:obj_removal}
3D object removal involves entirely eliminating the 3D Gaussians subset of an object from the scene. Results of such removals are illustrated in Figure~\ref{fig:3d_segmentation}. It's important to note that we apply a background bias $\gamma = -0.4$ across all scenes to ensure the background remains clear. For small 3D objects in row 1 and row 2, we simultaneously remove 5 objects from the scene. As the background of these two scenes is likely to be observed in other views, the artifacts in these two scenes are minor even imperceptible for certain objects. For more challenging scenes depicted in rows 3, 4, and 5, the space vacated by the removed objects reveals a noisier background or even black holes, primarily due to the obstruction of background by larger foreground objects. This situation is pronounced in the Horns scene (row 3), which comprises only facing-forward views.

\subsection{Object Inpainting} 

\begin{figure}[h!]
\centering
\includegraphics[width=1.0\linewidth]{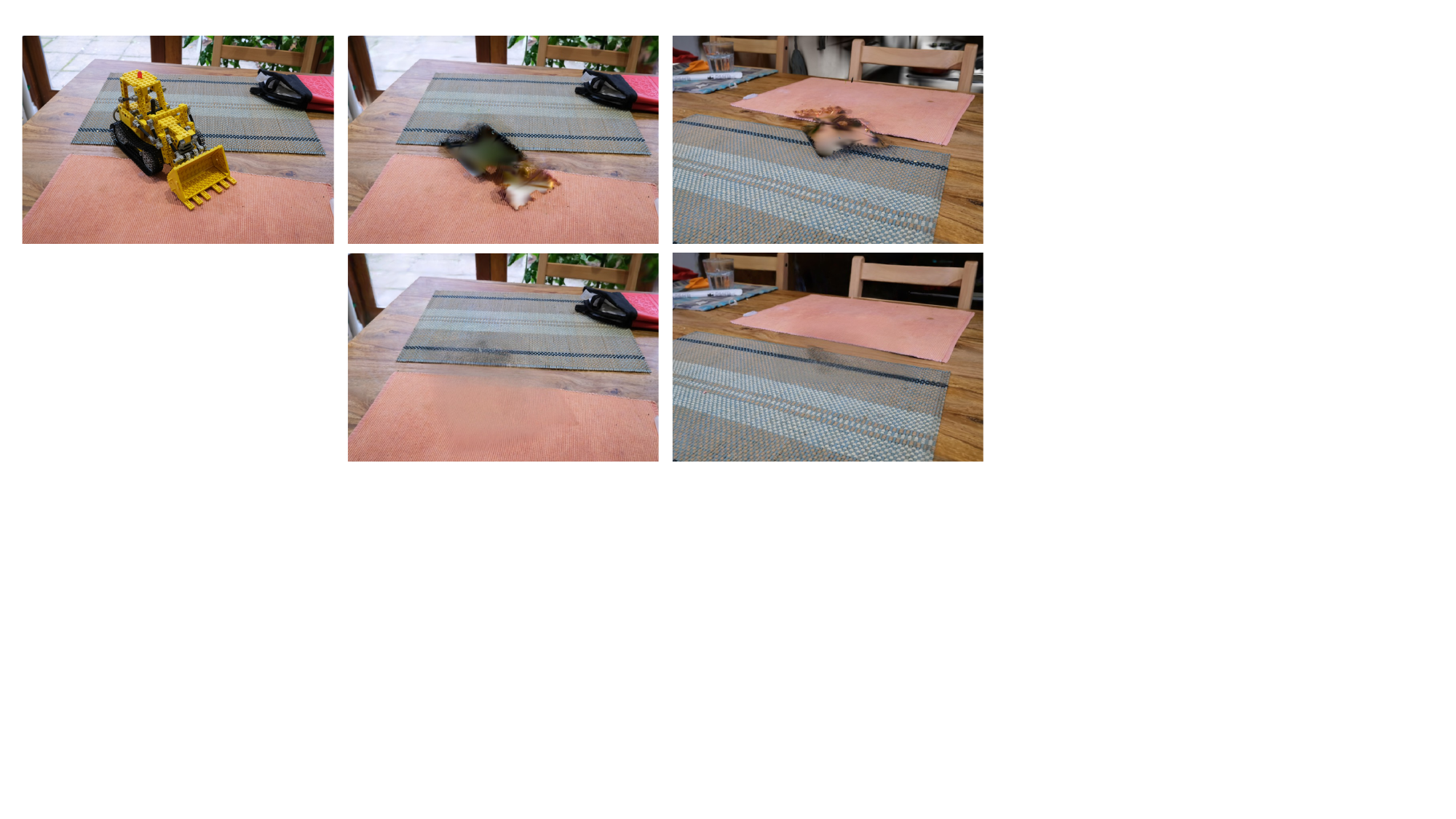}
\caption{\textbf{Object inpainting after removal}. Here we showcase the object removal results from 3D scenes, including the kitchen scene from MIP-360 dataset and bear scene from Instruct-NeRF2NeRF~\cite{instance-nerf}. After removal, we inpaint the regions with artifacts by tuning the optimized 3DGS parameters. Artifacts after object removal are diminished by our object inpainting.}
\label{fig:obj_removal_inpaint}
\vspace{-6mm}
\end{figure}

Following 3D object removal, object inpainting~\cite{qiu2021scene} aims to correct unobserved region artifacts, ensuring view consistency within the 3D scene. Initially, we render views post-removal and employ Grounding-DINO~\cite{ground-dino} to identify artifact regions in each view, which are tracked across views using a video tracker~\cite{track-anything}. Pretrained 2D inpainting models~\cite{lama} then generate inpainted 2D views. Subsequently, the 3DGS parameters are refined with these views by introducing $200K$ new Gaussians near the original object locations, maintaining background Gaussians frozen. Fine-tuning involves $L1$ loss outside object masks to minimize background impact, and LPIPS loss~\cite{lpips} inside the inpainting masks for scene naturalness and consistency. We display object inpainting results in Fig.~\ref{fig:obj_removal_inpaint}, rendering three views per scene. Noises and holes are diminished after the object inpainting, demonstrating that our method can effectively separate the foreground from the background in 3D segmentation.

\subsection{Quantitative comparison} 
We perform a quantitative analysis using the NVOS dataset~\cite{nvos}, derived from the LLFF dataset~\cite{llff}. The NVOS dataset comprises 8 facing-forward 3D scenes from LLFF, providing a reference and a target view with 2D segmentation mask annotations for each scene. Our approach begins with sampling point prompts from the reference view's annotated mask, which are then propagated to other views to guide the segmentation of SAM~\cite{sam}. Following this, we apply our binary segmentation approach as depicted in Eq.~\ref{eq:optimal_obj} to isolate the foreground 3D Gaussians. Subsequently, we employ novel view mask rendering with a threshold $\tau = 0.1$ to render 2D masks for the target view, allowing us to calculate the mean Intersection over Union (IoU) and mean accuracy across all 8 scenes. The comparative results, presented in Table~\ref{tab:nvos_eval}, benchmark our FlashSplat against previous NeRF-based 3D segmentation approaches such as NVOS~\cite{nvos}, ISRF~\cite{isrf}, SGIRF~\cite{sgisrf}, SA3D~\cite{sa3d}, and the 3D-GS method SAGA~\cite{saga}. Our method demonstrates superior performance in both metrics on this dataset.

\begin{table}[t]
\vspace{-4mm}
\centering
\begin{tabular}{c|cc}
\hline
Method            & mIOU ($ \%$) $\uparrow$     & mAcc ($ \%$) $\uparrow$     \\ \hline
NVOS~\cite{nvos}              & 39.4                        & 73.6                        \\
ISRF~\cite{isrf}              & 70.1                        & 92.0                        \\
SGISRF~\cite{sgisrf}            & 83.8                        & 96.4                        \\
SA3D~\cite{sa3d}              & 90.3                        & 98.2                        \\
SAGA~\cite{saga}              & {\color[HTML]{3531FF} 90.9} & {\color[HTML]{3531FF} 98.3} \\ \hline
FlashSplat (ours) & {\color[HTML]{FE0000} 91.8} & {\color[HTML]{FE0000} 98.6} \\ \hline
\end{tabular}
\caption{Quantitative comparison on the NVOS~\cite{nvos} dataset.}
\label{tab:nvos_eval}
\vspace{-6mm}
\end{table}

\subsection{Computation cost} 
We assess the computational efficiency of FlashSplat in comparison to previous 3DGS segmentation methods, namely SAGA~\cite{saga} and Gaussian Grouping~\cite{gsgroup}. This evaluation is conducted on the Figurines scene from the LERF dataset~\cite{lerf} using a single NVIDIA A6000 GPU. Both baseline methods necessitate the utilization of gradient descent optimization over $30,000$ iterations to distill 2D masks into object features associated with each 3D Gaussian, thereby incurring significant extra training time for optimized 3D scene. In contrast, our approach merely requires the calculation of the set $\{A_{i}\}_e$, a process that completes in approximately 26 seconds, making it roughly $50 \times$ faster than baselines. For segmenting a single 3D object, these baselines demand network forwarding, whereas FlashSplat efficiently employs $\argmax$ for optimal assignment determination, taking just $0.4$ milliseconds. Furthermore, an analysis of GPU memory usage reveals that our peak memory consumption is only half of that required by the previous method SAGA~\cite{saga}.

\begin{table}[]
\centering
\resizebox{0.85\linewidth}{!}{\begin{tabular}{c |C{1.8cm} C{1.9cm} C{1.8cm}  C{1.8cm}}
\hline
 & Extra Time & Optimization Steps & Segmentation Time & Peak Memory \\ \hline
SAGA~\cite{saga} & 18 min & 30000 & 0.5 s & 15G \\
Gaussian Grouping~\cite{gsgroup} & 37 min & 30000 & 0.3 s & 34G \\
\textbf{FlashSplat (ours)} & 26 s & 1 & 0.4 ms & 8G \\ \hline
\end{tabular}}
\caption{Computation cost comparisons over the Figurines scene.}
\label{tab:computation_cost}
\vspace{-6mm}
\end{table}


\vspace{-4mm}
\subsection{Ablation study}



\begin{figure}[h!]
\centering
\includegraphics[width=1.0\linewidth]{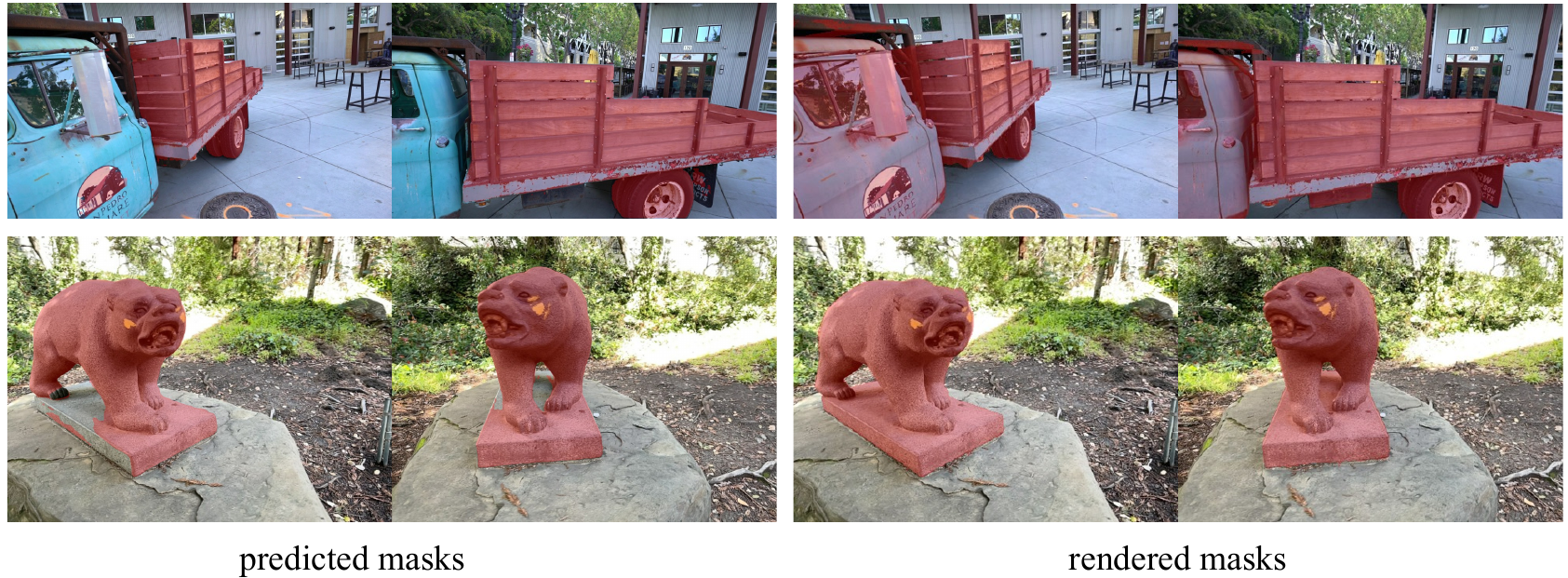}
\caption{\textbf{Noise reduction with 3D segmentation}. Here we show two noisy masks per scene, and the corresponding 2D mask is rendered from the 3D segmentation results. The broken regions in the given 2D masks are remedied from the 3D segmentation.}
\label{fig:noise_reduction}
\vspace{-4mm}
\end{figure}

\noindent{\textbf{The effect of noise reduction.}} To further elucidate the previously mentioned noise present in 2D masks, we provide visualizations of the 2D masks generated by SAM~\cite{sam} in the left column of Fig.~\ref{fig:noise_reduction} for two scenes. Additionally, we render the object mask after 3D segmentation on corresponding views, revealing that the broken regions in the provided 2D masks are remedied. This demonstrates the robustness of our method in generating 3D segmentation despite the presence of noise in the 2D masks.

\begin{figure}[h!]
\centering
\vspace{-4mm}
\includegraphics[width=1.0\linewidth]{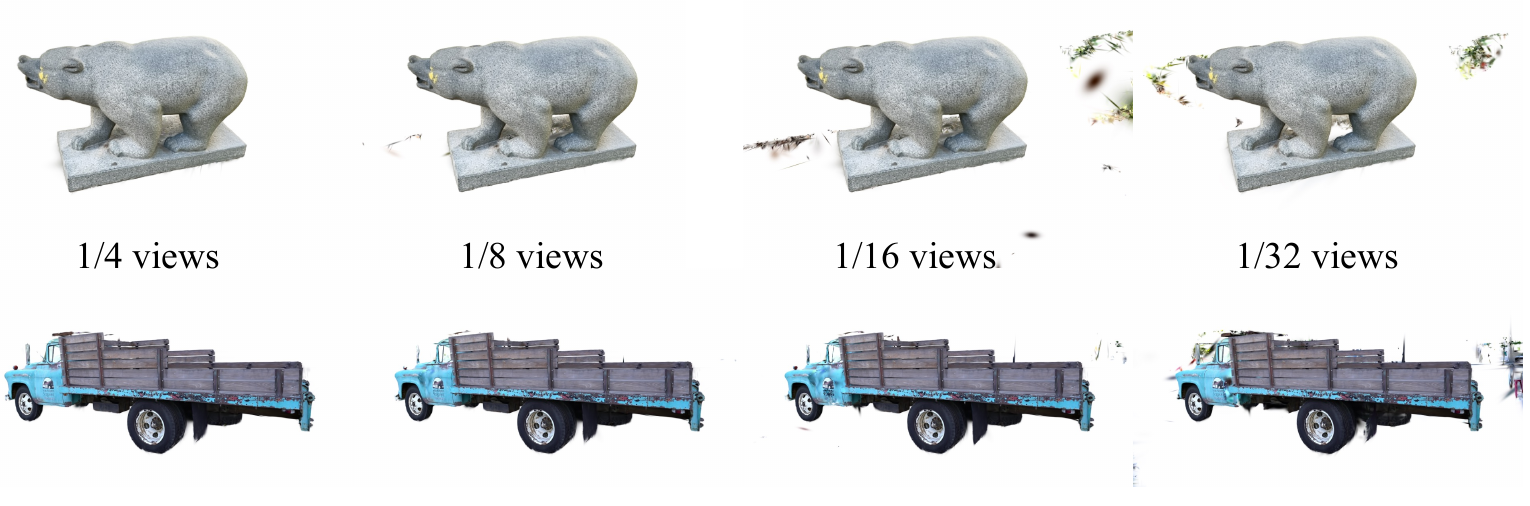}
\caption{\textbf{3D Segmentation with fewer masks}. Here we render 3D segmentation results on novel views with given fewer 2D masks, showing our method is capable of producing decent segmentation results with only $10\%$ of total view numbers.}
\label{fig:fewer_masks}
\vspace{-6mm}
\end{figure}

\noindent{\textbf{3D Segmentation with fewer 2D Masks.}} Since our approach does not rely on multi-step gradient descent optimization, it naturally enables the utilization of fewer 2D masks for segmentation. To validate this capability, we consider 2D mask subsets comprising $1/4$, $1/8$, $1/16$, and $1/32$ of the total number of views in the scene. Subsequently, we visualize the 3D segmentation results by rendering the foreground 3D Gaussians onto unseen views. As depicted in Fig.~\ref{fig:fewer_masks}, our method is capable of producing decent segmentation results even with only $1/8$ of the total view masks for these two 360-degree scenes. However, with fewer views, such as $1/16$ and $1/32$, many 3D Gaussians remain unseen in the provided masks, leading to potential artifacts in the segmentation outcomes.

\section{Conclusion}
In this work, we introduce an optimal solver for 3D Gaussian Splatting segmentation from 2D masks, significantly advancing the accuracy and efficiency of lifting 2D segmentation into 3D space. By breaking the alpha composition in 3D-GS into overall contributions of each Gaussians, this solver only requires single step optimization to get the optimal assignment. It not only expedit the optimization process by approximately 50 times faster compared to previous methods but also enhanced robustness against noise with a simple background bias. Further, this approach is extended to scene segmentation and capable of rendering masks on novel views. Extensive experiments demonstrate superior performance in scene segmentation tasks, including object removal and inpainting. We hope this work will facilitate 3D scene understanding and manipulating in the future. 

\section*{Acknowledgement}
This project is supported by the National Research Foundation, Singapore, under its Medium Sized Center for Advanced Robotics Technology Innovation.

\clearpage
\setcounter{page}{1}
\maketitlesupplementary

\section{More Implementation details}

The implementation of FlashSplat unfolds in two main parts. Initially, the focus is on deriving the contribution set $\{A_{e}\}$ for every Gaussian across objects $e$. This step involves projecting 3D Gaussians onto each mask $M_{v}$, capturing the product $\alpha_{i}T_{i}$ within the alpha blending formula into the buffer $A_{e}$ where a pixel $M_{ij}^{v} = e$. This procedure compiles the contributions from every object across all viewpoints into a matrix $\mathcal{A} \in \mathbb{R}^{E \times |\{G_{i}\}|}$, where $E$ is total number of objects in the 3D scene, and $|\{G_{i}\}|$ represents the total number of 3D Gaussians. Following this, we allocate labels $P_{i}$ to each 3D Gaussian $G_{i}$ based on this contribution matrix, as delineated in equations Eq.~\ref{eq:optimal_assignment} and Eq.~\ref{eq:multi_assignment}. For binary segmentation, the assignment process simplifies to an $\argmax$ operation for optimal assignment. Scene segmentation is resolved through dynamic programming to manage the complexity of multiple assignments, with the specific implementation details provided in list~\ref{lst:multi_instance_opt}. The segmentation results for scenes are represented within a matrix $\mathcal{S} \in \mathbb{R}^{E \times |\{G_{i}\}|}$, where each entry $\mathcal{S}_{m,n} \in \{0, 1\}$ specifies whether the $n$-th Gaussian belongs to object $m$.

\subsection{Mask association details}
Our work primarily concentrates on lifting 2D masks into 3D space, with less emphasis placed on mask association within the core sections of main paper. In this context, we provide additional insights into the methodology used to associate 2D masks in the 3D segmentation experiments presented.

\noindent{\textbf{Binary Mask Association.}} Within binary segmentation scenarios, association among 2D view masks is achieved through the propagation of point prompts across different views. Specifically, for a point prompt $p_{i}^{2D} \in \mathbb{R}^{2}$ identified on an object in a single view, this point is back-projected to the 3D space, acquiring a world coordinate $p_{i}^{3D} \in \mathbb{R}^{3}$, to locate its corresponding 3D Gaussian $G_{i}$. However, due to the prevalence of numerous Gaussians surrounding this 3D point prompt, relying solely on distance for Gaussian correspondence can lead to incorrect outcomes. To mitigate this, we initially identify the $\text{Top}-10$ closest 3D Gaussian centers using the $\mathcal{L}_2$ distance. Subsequently, the specific Gaussian is determined by selecting the one with the least depth when projected onto the reference view. The center positions of these 3D Gaussians are then projected onto other views to link point prompts associated with the same object across different views. By utilizing SAM~\cite{sam} to produce masks for each view based on these aligned point prompts, we inherently associate these predicted 2D masks. This method of point prompt propagation is implemented via CUDA kernels, enabling the association of point prompts across all views in under one second.

\noindent{\textbf{Scene Mask Association.}} Beyond binary segmentation, the segmentation of entire 3D scenes without specific point prompts is essential. Therefore, we introduce an alternative approach for associating scene masks across all views.  To assign each 2D object with a unique ID in the 3D scene, multiple views are treated akin to a video sequence. Utilizing a zero-shot video tracker~\cite{track-anything, gsgroup}, we ensure the consistent association and propagation of objects across viewpoints.

\section{More qualitative evaluation} 
\begin{figure}[h!]
\centering
\includegraphics[width=1.0\linewidth]{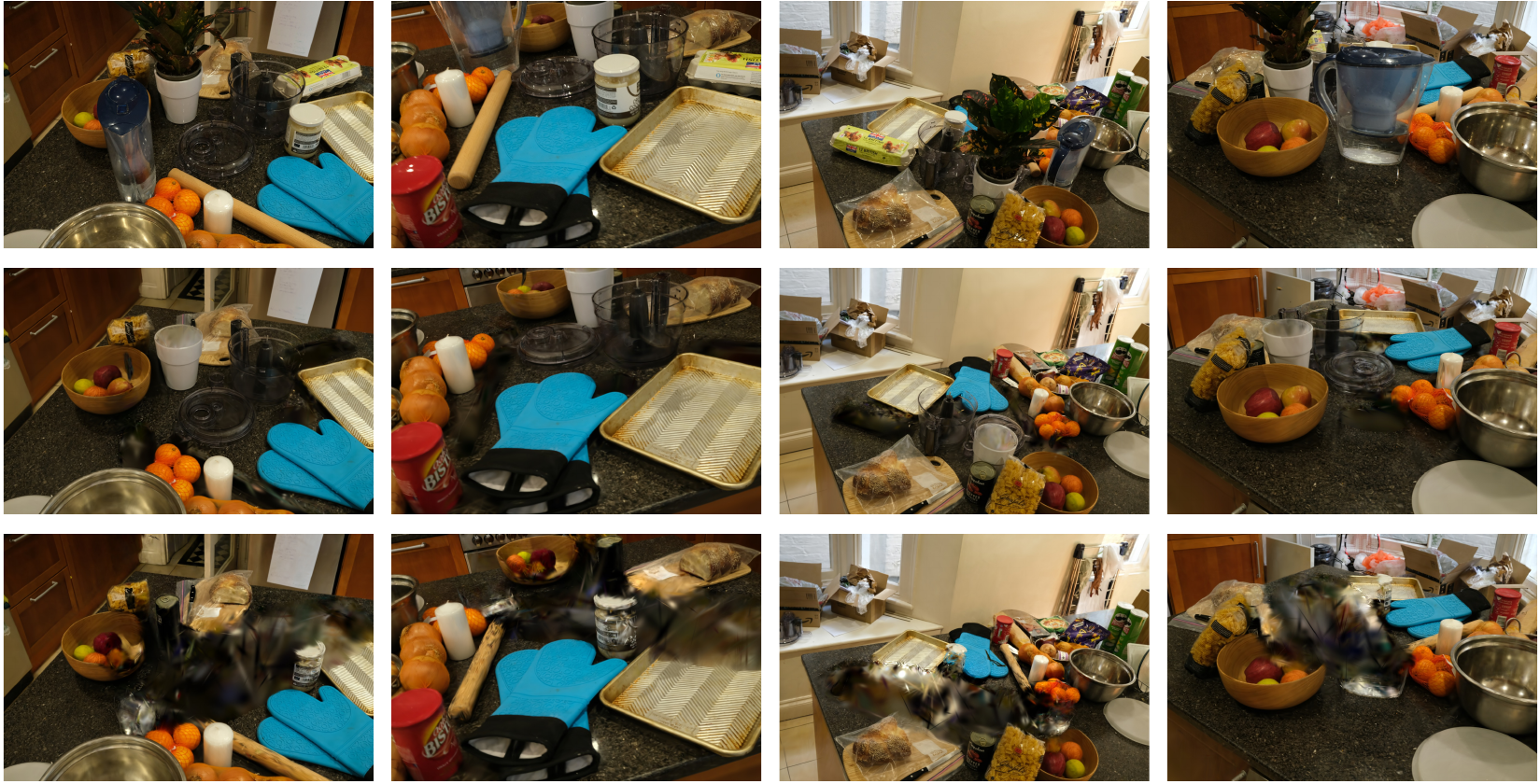}
\caption{\textbf{Multiple Object Removal Comparison}. Here we show a qualitative comparison by removing multiple objects from the Counter scene in the MIP-360~\cite{mipnerf360} dataset. The first row is the ground truth, the second row shows our FlashSplat, and the third row displays the results from Gaussian-Grouping. A total of 5 objects are removed in this scene, the same as in Fig.~\ref{fig:3d_segmentation} row 2.}
\label{fig:multi_removal}
\vspace{-6mm}
\end{figure}

\begin{figure}[h!]
\centering
\includegraphics[width=1.0\linewidth]{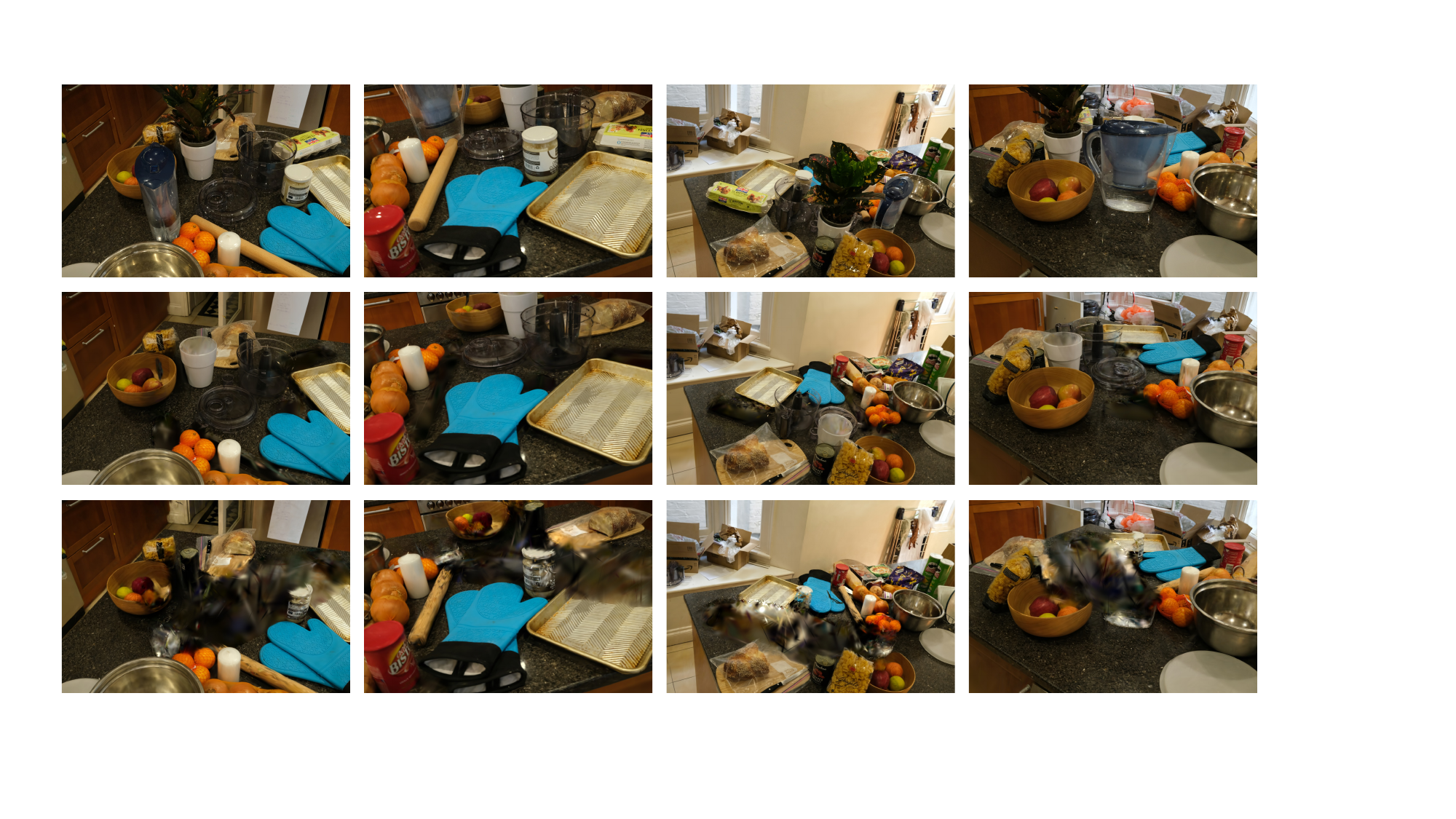}
\caption{\textbf{Single object removal comparison}. Here we show a qualitative comparison by removing a single object from the Bear scene in the Instruct-NeRF2NeRF~\cite{instance-nerf} dataset. The first row is the ground truth, the second row is our FlashSplat, and the third row shows the results from Gaussian-Grouping.}
\label{fig:single_removal}
\vspace{-4mm}
\end{figure}

To validate the effectiveness of our FlashSplat,  we conduct qualitative comparisons on the object removal task with prior works in 3D Gaussian Splatting segmentation, specifically Gaussian-Grouping~\cite{gsgroup}. As outlined in Sec.\ref{sec:obj_removal}, 3D object removal involves entirely eliminating the 3D Gaussians subset of selected objects from the scene, which is a fundamental application of 3D segmentation. For fair comparison, we use identical 2D scene mask set $\{M^v\}$ for both methods. Our process begins with conducting 3D segmentation using these scene masks, followed by the specification of object IDs for removal. We present the multiple object removal results in Fig.~\ref{fig:multi_removal} and single object removal results in Fig.~\ref{fig:single_removal}. We render 4 distinct views of the removal results, showing that our FlashSplat can cleanly remove these 3D objects with imperceptible artifacts, while the results of Gaussian-grouping show severe artifacts near the removed 3D objects. These comparisons underscore our method is not only superior for the efficiency of 3D segmentation, but also excels at 3D scene segmentation quality.

\section{More discussions}

\subsection{The effect of background bias $\gamma$}
Table~\ref{tab:softening_factor} presents our ablation study on the truck scenes from the T\&T dataset~\cite{tanks}. We annotate 5 views 2D mask as target views, and other view masks predicted by SAM~\cite{sam} are used as reference view masks. With the background bias $\gamma$ ranging from $[-1, 1]$, we get the 3D segmentation of the truck and then render it to 2D masks to compute the mean IoU. Among these $\gamma$ values, a setting of $\gamma = 0.4$ produces the optimal mean IoU of $94.2\%$. This is caused by the noise in masks predicted by SAM (as visualized in Fig.~\ref{fig:noise_reduction}), the assignment with $\gamma = 0$ is prone to take background Gaussians as foreground, while this softened refinement helps to reduce such noises. 

\begin{table}[h!]
\vspace{-4mm}
\centering
\setlength{\tabcolsep}{16pt} 
\begin{tabular}{c|ccccc }
\hline
$\gamma$ & -0.8 & -0.4 & 0    & 0.4  & 0.8  \\ \hline
mIoU     & 82.4 & 89.6 & 92.3 & 94.2 & 93.8 \\ \hline
\end{tabular}
\caption{Effect of the background bias $\gamma$ on the truck scene.}
\label{tab:softening_factor}
\vspace{-6mm}
\end{table}

\subsection{Quantization in novel view mask}
\begin{figure}[h!]
\centering
\includegraphics[width=1.0\linewidth]{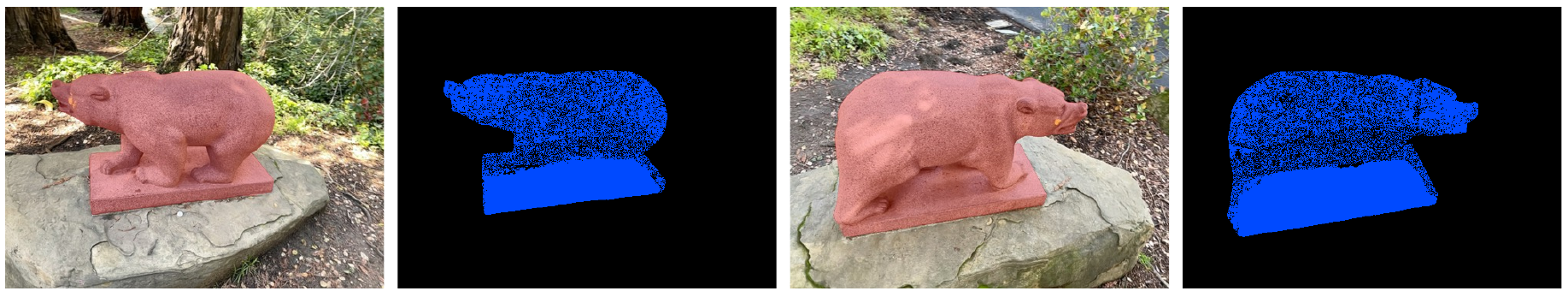}
\caption{\textbf{Quantization in mask rendering}.}
\label{fig:quant_mask}
\end{figure}

In Sec.~\ref{sec:mask_render}, we outline projecting masks from 3D segmentation results onto novel views using simple quantization and depth-guidance. Here we take mask rendering in binary segmentation as an example to claim why this quantization is necessary. Without this quantization, novel view masks are produced by first projecting subsets of 3D Gaussians $\{G_{i}\}_{e}$ for instance $e$, and then the mask value for each pixel $\hat{M}^{v}_{jk}$ is determined by $\hat{M}^{v}_{jk} = \argmax_{e}{\rho_{jk}^{e}}$. The absence of quantization leads to masks, displayed in the $2$-nd column and $4$-th row of Fig.~\ref{fig:quant_mask}, riddled with numerous unintended holes. This phenomenon stems from the nature of 3D Gaussian Splatting, where each Gaussian is a semi-transparent ellipse with opacity $o_i$. As such, when a pixel $(i, j)$ is rendered, the background 3D Gaussians also affect the alpha blending outlined in Eq.~\ref{eq:alpha_blend}, at times more significantly than the foreground Gaussians. Implementing quantization and depth guidance ameliorates these discrepancies, as evident in the first and third columns of Fig.~\ref{fig:3d_segmentation}. However, it's noteworthy that despite the employment of depth guidance, mask rendering can still produce vague outcomes in scene mask rendering due to the absence of geometric supervision in 3D Gaussian Splatting reconstruction. The geometry that is learned does not conform precisely to the underlying geometry, occasionally impairing the effectiveness of depth guidance.

\subsection{Scene segmentation extension}
\label{sec:mext_dicussion}
\begin{figure}[h!]
\setlength{\abovecaptionskip}{0cm}
\setlength{\belowcaptionskip}{0cm}
\vspace{-4mm}
\centering
\includegraphics[width=0.8\linewidth]{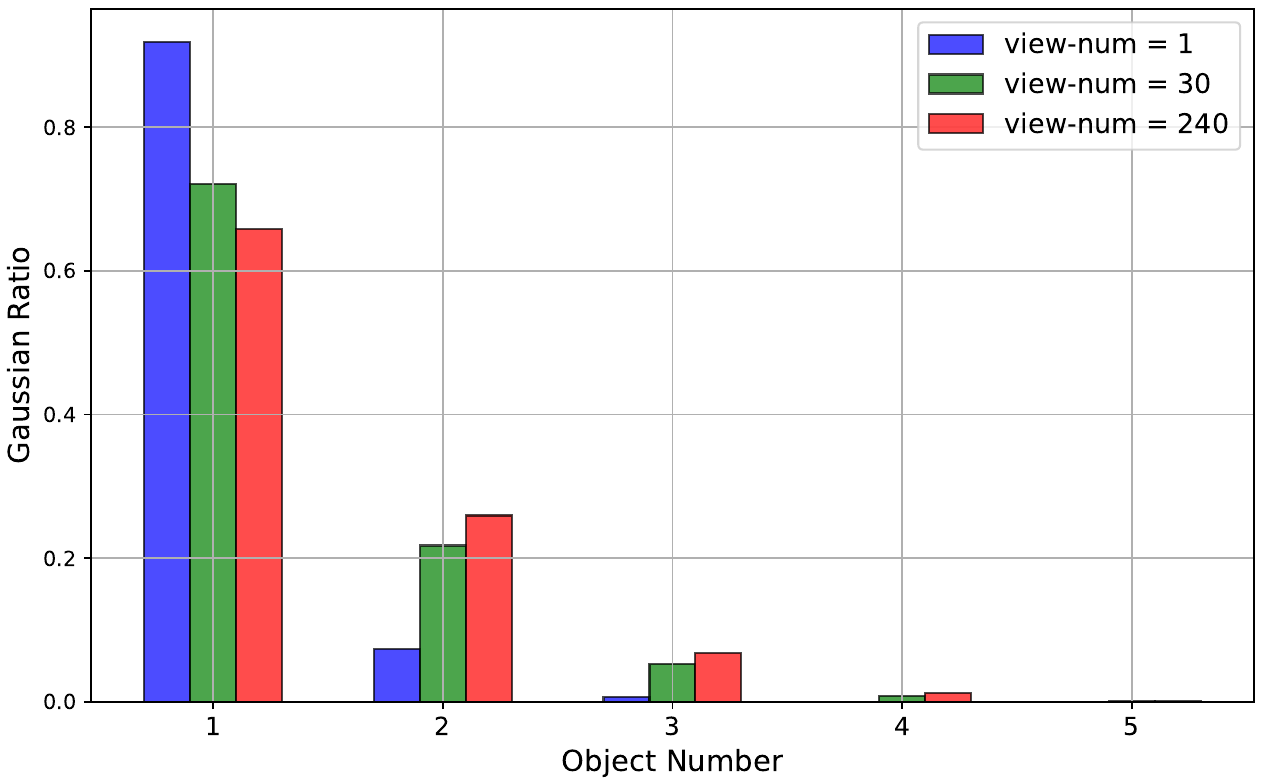}
\caption{Visualization of each Gaussian's contributed object number.}
\label{fig:ablation_nonexclu}
\vspace{-4mm}
\end{figure}

In Sec.~\ref{subsec:multi-seg}, we extend our optimal assignment for binary segmentation to scene segmentation. This formulation, shown in Eq.~\ref{eq:multi_assignment}, is chosen over a straightforward approach that would simply perform $\argmax$ among the $E$ instances. This choice is driven by the non-exclusive nature of Gaussian Splatting, where a Gaussian can be shared between objects. For instance, we quantitatively analyze the Counter scene in the MIP360~\cite{mipnerf360} dataset under different numbers of given masks. As illustrated in Fig.~\ref{fig:ablation_nonexclu}, approximately $20\%$ of the Gaussians in this scene are shared between more than two objects. This phenomenon occurs because, in the 3D reconstruction of 3DGS, supervision is limited to view space, with no additional geometric or semantic constraints to enforce mutual exclusivity.

\section{Limitations}
Despite the advancements presented by our method in 3D-GS segmentation, we acknowledge several limitations for future exploration. The linear programming approach, although effective, may encounter scalability challenges with significantly larger 3D scenes with substantial spatial resolutions, as we need to traverse all mask pixels. Moreover, due to the inherent property of 3D-GS, rendering 3D segmentation results onto novel view with depth guidance may currently yield ambiguous mask. To address this limitation,  incorporating explicit geometry supervision into the 3D-GS reconstruction is essential for more accurately representing the underlying geometry. Additionally, the investigation of adaptive strategies aimed at reducing computational demands and enhancing the adaptability of our method to handle a broader array of scene complexities presents a promising future work.

\newpage
\begin{lstlisting}[caption={Multi-instance optimization in PyTorch}, label=lst:multi_instance_opt, float]
def multi_instance_opt(all_contrib, gamma=0.):
    """
    Input:
    all_contrib: A_{e} with shape (obj_num, gs_num) 
    gamma: background bias range from [-1, 1]
    
    Output: 
    all_obj_labels: results S with shape (obj_num, gs_num)
    where S_{i,j} = 1 denotes j-th gaussian belong i-th object
    """
    all_contrib_sum = all_contrib.sum(dim=0)
    all_obj_labels = torch.zeros_like(all_contrib)
    for obj_idx, obj_contrib in enumerate(all_contrib):
        other_contrib = all_contrib_sum - obj_contrib
        obj_contrib = torch.stack([other_contrib, obj_contrib])
        obj_contrib = F.normalize(obj_contrib, dim=0, p=1)
        obj_contrib[0, :] += gamma
        obj_label = torch.argmax(obj_contrib, dim=0)
        all_obj_labels[obj_idx] = obj_label
    return all_obj_labels
\end{lstlisting}


\clearpage  



\bibliographystyle{splncs04}
\bibliography{main}

\end{document}


\title{FlashSplat: 2D to 3D Gaussian Splatting Segmentation Solved Optimally} 

\titlerunning{Abbreviated paper title}

\author{First Author\inst{1}\orcidlink{0000-1111-2222-3333} \and
Second Author\inst{2,3}\orcidlink{1111-2222-3333-4444} \and
Third Author\inst{3}\orcidlink{2222--3333-4444-5555}}

\authorrunning{F.~Author et al.}

\institute{Princeton University, Princeton NJ 08544, USA \and
Springer Heidelberg, Tiergartenstr.~17, 69121 Heidelberg, Germany
\email{lncs@springer.com}\\
\url{http://www.springer.com/gp/computer-science/lncs} \and
ABC Institute, Rupert-Karls-University Heidelberg, Heidelberg, Germany\\
\email{\{abc,lncs\}@uni-heidelberg.de}}


\maketitlesupplementary

\section{introduction}

\clearpage  

%
%
\bibliographystyle{splncs04}
\bibliography{main}